\documentclass[preprint,12pt]{elsarticle}

\usepackage[T1]{fontenc}
\usepackage[utf8]{inputenc}
\usepackage{microtype}
\usepackage{graphicx}
\usepackage{amsmath,amssymb,amsfonts}
\usepackage{amsthm}
\usepackage{booktabs}
\usepackage{tabularx}
\usepackage{array}
\usepackage{xcolor}
\usepackage{algorithm}
\usepackage{algorithmicx}
\usepackage{algpseudocode}
\usepackage{subcaption}
\usepackage{enumitem}
\usepackage{tikz}
\usepackage{pgfplots}
\usetikzlibrary{positioning,fit,backgrounds,arrows.meta,calc}
\pgfplotsset{compat=1.18}
\usepackage{hyperref}
\hypersetup{
    hidelinks,
    pdftitle={When Explanations Compete: Policy-Aware Selection Under Uncertainty},
    pdfauthor={Helena Löfström, Tuwe Löfström, Johan Hallberg Szabadváry}
}

\journal{Information Sciences}
\graphicspath{{figures/}}

\newcolumntype{Y}{>{\raggedright\arraybackslash}X}
\newcolumntype{L}[1]{>{\raggedright\arraybackslash}p{#1}}
\newcolumntype{C}[1]{>{\centering\arraybackslash}p{#1}}

\newcommand{\CalA}{\mathcal{A}}
\newcommand{\Aelig}{\CalA_{\mathrm{elig}}}

\theoremstyle{plain}
\newtheorem{proposition}{Proposition}[section]

\author[jibs]{Helena Löfström\corref{cor1}}
\ead{helena.lofstrom@ju.se}
\author[jth]{Tuwe Löfström}
\author[jth]{Johan Hallberg Szabadváry}
\cortext[cor1]{Corresponding author.}

\affiliation[jibs]{
    organization={Jönköping International Business School},
    addressline={Gjuterigatan 5},
    city={Jönköping},
    postcode={553 18},
    country={Sweden}
}
\affiliation[jth]{
    organization={Jönköping School of Engineering},
    addressline={Gjuterigatan 5},
    city={Jönköping},
    postcode={553 18},
    country={Sweden}
}

\begin{document}
\begin{frontmatter}

\title{When Explanations Compete: Policy-Aware Selection Under Uncertainty}

\begin{abstract}
Uncertainty-aware explanation methods often produce several alternatives for the same prediction. Selecting among them requires a policy for balancing prediction confidence, uncertainty, and application constraints. This paper presents a framework for applying such policies to a fixed set of generated explanations. Candidates are characterised by uncertainty change, prediction direction, and, when available, interval position relative to a decision boundary. The framework combines these properties with eligibility rules, optional bidirectional Pareto screening, and policy-aware ranking. A fictitious prostate-cancer example illustrates how different explanatory purposes lead to different selections from the same candidate set. We instantiate the framework with Calibrated Explanations for classification, thresholded regression, and plain regression. Across 41 benchmark datasets, mean candidate counts range from 11.57 to 21.75 for single-feature explanations and from $29.48$ to $69.53$ when conjunctions are included. Equal-weight and confidence-only policies yield an average selection-disagreement rate of $28.7\%$ while favouring the same confidence direction. A supporting $\delta$-CLUE experiment demonstrates use with a second generator. By making the selection policy explicit, the framework allows applications to compare and prioritise explanations according to their intended use.
\end{abstract}


\end{frontmatter}

\section{Introduction}\label{sec:introduction}
An explanation generator may return several plausible alternatives for the same prediction. A decision-support system must still determine which alternatives to show, compare, or prioritise. This choice becomes consequential when candidates differ in both prediction and uncertainty: a large change in a decision-relevant prediction may be less well supported than a smaller change, while the lowest-uncertainty alternative may answer the wrong explanatory question. Candidate generation therefore leaves a selection problem whose answer depends on the intended use of the explanation.

Alternative explanations make this problem particularly visible. Counterfactual methods search for changes associated with a different prediction, often balancing proximity, plausibility, diversity, and actionability \cite{wachter2017counterfactual,mothilal2020explaining,karimi2021algorithmic}. Semifactual explanations retain an outcome under changed conditions, and explanation-set approaches organise several alternative forms through explicit restrictions \cite{mccloy2002semifactual,fernandez2022explanation}. Uncertainty-aware methods additionally communicate reliability information or search for lower-uncertainty alternatives \cite{bhatt2021uncertainty,antoran2020getting,ley2021delta}. Once several candidates are available, however, neither validity nor low uncertainty alone establishes which candidate is most appropriate.

Consider a fictitious prostate-cancer decision-support case. A 72-year-old patient has prostate-specific antigen (PSA) of 4.2 ng/ml, prostate volume of 28 ml, and PI-RADS score 3. We use this context only to make the example concrete. All model outputs, intervals, thresholds, and alternatives are invented and have no intended clinical interpretation. Let $C(x)=0.40$ denote the decision-relevant model confidence in clinically significant cancer for the factual case, with interval $[0.20,0.60]$ and illustrative decision threshold $\tau=0.35$. The interval spans the threshold, so a point estimate alone leaves relevant ambiguity.

Suppose one alternative yields confidence 0.25 with interval $[0.20,0.30]$, while another yields confidence 0.85 with interval $[0.80,0.90]$. Both reduce interval width equally, yet the first challenges the factual threshold classification and the second reinforces it. A third alternative reduces uncertainty but changes an immutable feature, while a fourth offers higher confidence at the cost of greater uncertainty. Which alternative should be prioritised depends on the explanatory purpose and the application's constraints. Making these choices explicit allows the selected explanation to be interpreted in relation to the question it answers.

Post-generation selection has been studied by St\k{e}pka et al.~\cite{stepka2024multi}, who combine counterfactual generators with constraints, dominance filtering, and multi-criteria selection. Explanation Sets \cite{fernandez2022explanation} unifies counterfactual and semifactual explanations through explicit restrictions. Building on these approaches, we formalise how uncertainty change, boundary ambiguity, and prediction direction inform selection under explicit application policies.

The main contribution is a reusable framework for applying explicit selection policies to uncertainty-aware explanations. It separates three questions: what properties a candidate has, whether it satisfies the application's requirements, and how it should be prioritised for the current explanatory purpose. The same generated candidate set can therefore support different explanatory questions and policy preferences. The framework specifies the information required for each operation, the treatment of unavailable information, and the selection results preserved by optional screening. A worked example, formal analysis, and experiments demonstrate complementary aspects of this framework.

We instantiate the framework with Calibrated Explanations (CE) for binary and multiclass classification, thresholded regression, and plain regression. Across 41 benchmark datasets, the generated sets contain multiple alternatives, including candidates whose uncertainty is no higher than that of the factual case. The policy comparison shows that incorporating uncertainty changes which explanation is selected even when the preferred confidence direction remains fixed.

The remainder of the paper positions the contribution against related work, defines the framework and worked example, maps CE outputs to the interface, reports the empirical diagnostics, and discusses limitations and deployment implications.

\section{Related Work}\label{sec:related-work}

\paragraph{Alternative explanations, semifactuals, and recourse.}
Counterfactual explanations describe changes associated with a different model output \cite{wachter2017counterfactual,mothilal2020explaining,guidotti2022counterfactual,verma2024counterfactual}. Semifactual explanations describe changes under which an outcome persists \cite{mccloy2002semifactual}. Recourse adds requirements concerning feasible interventions and their consequences \cite{karimi2021algorithmic}. These notions should not be conflated. A candidate may reveal model behaviour without describing an action that a person can or should take. Explanation Sets \cite{fernandez2022explanation} is especially relevant because it offers a general framework for counterfactual and semifactual explanations, supports restrictions, and extends to regression. Our directional labels do not replace those categories. They describe how a candidate point prediction moves relative to a supplied boundary, while uncertainty and eligibility remain separate axes. Conventional semifactuals are broader than the semi-oriented subset defined below: a semifactual can preserve an outcome while moving toward, away from, or neither relative to a chosen boundary.

\paragraph{Uncertainty-aware explanation generation.}
Predictive uncertainty and uncertainty in an explanation are related but distinct concerns. Prior work motivates uncertainty communication and attaches uncertainty information to predictions or post-hoc explanations \cite{hullermeier2021aleatoric,bhatt2021uncertainty,tomsett2020rapid,slack2021reliable,marx2023but}. Conformal and calibration-based approaches provide further uncertainty-aware explanatory representations \cite{narteni2023confiderai,mehdiyev2025augmenting}. CLUE searches for nearby alternatives with lower predictive uncertainty, while $\delta$-CLUE and DIVERSE CLUE expose multiple alternatives \cite{antoran2020getting,ley2021delta,ley2022diverse}. QUCE incorporates uncertainty associated with a counterfactual path \cite{duell2024quce}. These methods demonstrate that uncertainty-aware generation is feasible, but the uncertainty preference is often coupled to the generator's objective or search procedure. When several candidates remain, an application still needs to decide how uncertainty, direction, and eligibility affect presentation.

\paragraph{Multi-objective generation and post-generation selection.}
Multi-objective counterfactual methods optimise competing criteria such as validity, proximity, sparsity, feasibility, and diversity during search \cite{dandl2020multi}. St\k{e}pka et al.~\cite{stepka2024multi} address the later choice among generated candidates by combining counterfactuals from multiple explainers, enforcing constraints, removing dominated candidates, and applying an ideal-point selector. Their work is a direct precedent for post-generation selection. Our scope is narrower in optimisation terms but broader in explanatory typing: the framework keeps a candidate-level uncertainty signal, its change from the factual case, threshold ambiguity, confidence direction, and application eligibility explicit across classification and regression settings. The same fixed candidate set can then answer different uncertainty-aware explanatory questions without regeneration.

Table~\ref{tab:positioning} summarises the closest distinctions. The proposed interface is compatible with generators that already embed uncertainty or multiple objectives, but it cannot recover candidates excluded during generation and it does not make heterogeneous uncertainty measures semantically equivalent.

\begin{table}[t]
\centering
\footnotesize
\caption{Positioning against closely related approaches.}
\label{tab:positioning}
\begin{tabularx}{\linewidth}{L{2.9cm}L{2.5cm}Y}
\toprule
Approach & Primary selection stage & Relation to the present framework \\
\midrule
Explanation Sets \cite{fernandez2022explanation} & Representation and restriction & Unifies counterfactual and semifactual explanations and supports restrictions. Not organised around candidate-level uncertainty trade-offs. \\
CLUE family \cite{antoran2020getting,ley2021delta,ley2022diverse} & During generation & Searches for uncertainty-aware alternatives. Multiple returned candidates can still require an external policy. \\
St\k{e}pka et al.~\cite{stepka2024multi} & After generation & Closest selection precedent. Applies multi-criteria filtering and ideal-point selection to counterfactuals from an ensemble of explainers. \\
Proposed framework & After generation & Separates uncertainty change, threshold ambiguity, confidence direction, eligibility, and policy across compatible alternative-explanation records. \\
\bottomrule
\end{tabularx}
\end{table}

\section{Uncertainty-Aware Selection Framework}\label{sec:framework}
\subsection{Candidate records, factual reference, and policy}\label{subsec:interface}
Let $x$ be a factual case and $\CalA(x)=\{a_1,\ldots,a_m\}$ a finite candidate set returned by an explanation generator. We write $C(z)$ for the decision-relevant prediction coordinate of an instance $z$ and $U(z)$ for its uncertainty signal, where lower $U$ denotes less uncertainty under the chosen estimator. Each record $a_j$ identifies an alternative instance or rule description $x'_j$ and stores the shorthand coordinates $C_j:=C(x'_j)$ and $U_j:=U(x'_j)$. The corresponding factual coordinates are $C_0:=C(x)$ and $U_0:=U(x)$. Optional fields include interval bounds $[\ell_j,h_j]$, changed feature conditions, proximity, cost, feasibility, generator identity, and provenance.

Within a case, $C_j$ must refer to the same class, event, or numeric outcome across candidates. The uncertainty signal must likewise support a consistent within-case ordering, with lower values denoting less uncertainty under that estimator. Numerical scaling can facilitate ranking, but it cannot make entropy, interval width, path unreliability, or other quantities semantically equivalent. Assigning uncertainty-change status requires $U_0$. Assigning threshold ambiguity requires actual bounds and a relevant boundary. It cannot generally be inferred from a point estimate and width alone. Enforcing a domain restriction requires the corresponding metadata or an external eligibility predicate.

A selection policy specifies the explanatory direction, permitted candidate properties, eligibility constraints, uncertainty tolerance, ranking transformation, numerical preference, screening choice, tie rule, and requested output. Candidate properties remain separate from these choices. An immutable-feature alternative can be mathematically valid and informative about the model while being ineligible for actionable recourse. Missing metadata is treated as unavailable rather than silently interpreted as satisfying a constraint.

\begin{figure}[t]
\centering
\begin{tikzpicture}[
  node distance=6mm,
  box/.style={draw,rounded corners,align=center,text width=0.27\linewidth,minimum height=12mm,font=\small},
  accent/.style={box,fill=black!5,line width=0.7pt},
  arrow/.style={-{Latex[length=2mm]},thick}]
\node[box] (gen) {Candidate generator\\$\CalA(x)$};
\node[accent,right=of gen] (sel) {Selection contract\\properties, eligibility, policy};
\node[box,right=of sel] (out) {Selected record, shortlist,\\eligible set, or empty result};
\draw[arrow] (gen)--(sel);
\draw[arrow] (sel)--(out);
\end{tikzpicture}
\caption{The framework operates after generation. It preserves candidate records and makes application-specific selection choices explicit.}
\label{fig:pipeline}
\end{figure}

\subsection{Compositional candidate properties}\label{subsec:properties}
Three overlapping axes describe a candidate: uncertainty change, interval position relative to a boundary, and movement of the point prediction. They are compositional properties, not statistically independent dimensions. A candidate can, for example, be ensured, potential, and counter-oriented simultaneously.

\begin{table}[t]
\centering
\footnotesize
\caption{Operational candidate properties. ``Ensured'' concerns the selected uncertainty signal; it is not a guarantee of correctness, calibration after selection, feasibility, or actionability.}
\label{tab:properties}
\begin{tabularx}{\linewidth}{L{3.0cm}L{5.1cm}Y}
\toprule
Property & Condition & Interpretation \\
\midrule
Ensured & $U_j<U_0$. Optionally $U_j\leq U_0-\epsilon_U$ with $\epsilon_U>0$  & Strict reduction in the specified uncertainty signal. \\
Uncertainty-non-increasing & $U_j\leq U_0$ & Includes equal uncertainty, used by the CE diagnostic reported later. \\
Uncertainty-increasing & $U_j>U_0$ & Candidate carries greater uncertainty than the factual reference. \\
Potential & $\ell_j\leq\tau\leq h_j$ & Candidate interval spans or touches the supplied boundary. \\
Counter-oriented & $(C_j-\tau)(C_0-\tau)<0$ & Point prediction crosses to the opposite side of the boundary. \\
Semi-oriented & $(C_j-\tau)(C_0-\tau)>0$ and $|C_j-\tau|<|C_0-\tau|$ & Remains on the factual side but moves toward the boundary. \\
Super-oriented & $(C_j-\tau)(C_0-\tau)>0$ and $|C_j-\tau|>|C_0-\tau|$ & Remains on the factual side and moves away from the boundary. \\
\bottomrule
\end{tabularx}
\end{table}

Ensured status says only that $U_j$ is lower under the chosen operational measure. Potential status inherits the statistical meaning of the supplied interval. Directional labels describe point-prediction movement and do not establish causal effects. Equality cases remain explicit: an unchanged point prediction receives an unchanged label, and a point exactly at $\tau$ is treated as a boundary case rather than forced into a strict directional category. In plain regression without an application reference, direction is recorded as increase, decrease, or equality relative to $C_0$.

\subsection{Worked example: direction, uncertainty, and eligibility}\label{subsec:cancer}
Table~\ref{tab:cancer} extends the fictitious cancer example to four alternatives. The factual interval has width $U_0=0.40$. The candidates are deliberately constructed to isolate three distinct questions: opposite confidence directions at equal uncertainty, threshold ambiguity and eligibility, and a same-direction confidence--uncertainty trade-off.

\begin{table}[t]
\centering
\footnotesize
\setlength{\tabcolsep}{3.4pt}
\caption{Fictitious prostate-cancer alternatives. The factual record is $C_0=0.40$, interval $[0.20,0.60]$, $U_0=0.40$, and $\tau=0.35$. All predictive quantities and alternatives are illustrative.}
\label{tab:cancer}
\begin{tabularx}{\linewidth}{C{0.4cm}Y C{0.60cm}C{1.65cm}C{0.60cm}L{4.5cm}}
\toprule
 & Hypothetical alternative & $C_j$ & Interval & $U_j$ & Properties and possible eligibility \\
\midrule
$a_1$ & Refined prostate-volume measurement yields PSA density $<0.10$ & 0.25 & $[0.20,0.30]$ & 0.10 & Ensured and counter-oriented: interval wholly below $\tau$. \\
$a_2$ & Patient's age is 65 years & 0.30 & $[0.15,0.45]$ & 0.30 & Ensured, potential, and counter-oriented: excluded if age changes are prohibited. \\
$a_3$ & MRI re-evaluation yields PI-RADS 5 & 0.85 & $[0.80,0.90]$ & 0.10 & Ensured and super-oriented: interval wholly above $\tau$. \\
$a_4$ & PI-RADS 5 together with PSA density $>0.20$ & 0.90 & $[0.80,1.00]$ & 0.20 & Ensured and super-oriented: higher confidence but greater uncertainty than $a_3$. \\
\bottomrule
\end{tabularx}
\end{table}

\begin{figure}[t]
\centering
\begin{subfigure}[t]{0.53\linewidth}
\centering
\begin{tikzpicture}[x=5.85cm,y=0.61cm,font=\scriptsize]
  \draw[->] (0,0.45) -- (1.04,0.45) node[right] {$C$};
  \foreach \x in {0,0.2,0.4,0.6,0.8,1.0}{
    \draw (\x,0.39)--(\x,0.51);
    \node[below] at (\x,0.39) {\x};
  }
  \draw[dashed,thick] (0.35,0.45)--(0.35,5.55);
  \node[above,rotate=90] at (0.35,5.55) {$\tau=0.35$};
  \foreach \lab/\y in {$x$/5,$a_1$/4,$a_2$/3,$a_3$/2,$a_4$/1}{\node[left] at (0,\y) {\lab};}
  \draw[line width=1.2pt] (0.20,5)--(0.60,5); \filldraw (0.40,5) circle (1.7pt);
  \draw[line width=1.2pt] (0.20,4)--(0.30,4); \filldraw (0.25,4) circle (1.7pt);
  \draw[line width=1.2pt] (0.15,3)--(0.45,3); \filldraw (0.30,3) circle (1.7pt);
  \draw[line width=1.2pt] (0.80,2)--(0.90,2); \filldraw (0.85,2) circle (1.7pt);
  \draw[line width=1.2pt] (0.80,1)--(1.00,1); \filldraw (0.90,1) circle (1.7pt);
\end{tikzpicture}
\caption{Point predictions and intervals.}
\label{fig:cancer-intervals}
\end{subfigure}
\hfill
\begin{subfigure}[t]{0.43\linewidth}
\centering
\begin{tikzpicture}[x=4.38cm,y=8.2cm,font=\scriptsize]
  \draw[->] (0,0)--(1.04,0) node[right] {$C$};
  \draw[->] (0,0)--(0,0.46) node[above] {$U$};
  \foreach \x in {0,0.2,0.4,0.6,0.8,1.0}{\draw (\x,-0.008)--(\x,0.008); \node[below] at (\x,-0.008) {\x};}
  \foreach \y in {0.1,0.2,0.3,0.4}{\draw (-0.01,\y)--(0.01,\y); \node[left] at (-0.01,\y) {\y};}
  \coordinate (f) at (0.40,0.40);
  \coordinate (a1) at (0.25,0.10); \coordinate (a2) at (0.30,0.30);
  \coordinate (a3) at (0.85,0.10); \coordinate (a4) at (0.90,0.20);
  \foreach \p in {a1,a2,a3,a4}{\draw[gray,densely dotted] (f)--(\p);}
  \filldraw[black] (f) rectangle ++(0.022,0.013); \node[above left] at (f) {$x$};
  \filldraw[blue!70!black] (a1) circle (2pt); \node[below left] at (a1) {$a_1$};
  \filldraw[orange!85!black] (a2) circle (2pt); \node[above left] at (a2) {$a_2$};
  \filldraw[green!45!black] (a3) circle (2pt); \node[below left] at (a3) {$a_3$};
  \filldraw[purple!75!black] (a4) circle (2pt); \node[above right] at (a4) {$a_4$};
\end{tikzpicture}
\caption{The same records in confidence--uncertainty space.}
\label{fig:cancer-space}
\end{subfigure}
\caption{The worked example separates direction, uncertainty, and eligibility. Candidates $a_1$ and $a_3$ have equal uncertainty but opposite confidence directions; $a_3$ and $a_4$ move in the same direction but trade confidence against uncertainty.}
\label{fig:cancer}
\end{figure}

Candidate $a_1$ challenges the factual threshold classification, whereas $a_3$ reinforces it at the same uncertainty. Candidate $a_2$ shows why candidate type and eligibility must remain separate: it satisfies the mathematical properties but can be excluded by a policy prohibiting changes to age. Candidate $a_4$ is still ensured relative to the factual case, yet it is more uncertain than $a_3$. Thus, uncertainty reduction relative to the factual case does not uniquely order ensured alternatives.

The example uses interval midpoints as the generic confidence coordinate. This convention is chosen only to make the geometry transparent. It is not a claim that CE's Venn--Abers point output is always the midpoint of its probability pair. Under the single-pair transformation $C=h/(1-\ell+h)$, the $a_3$ and $a_4$ point values and their precise switching weight would differ, while the same-direction trade-off would remain.

\subsection{Filtering, screening, ranking, and outputs}\label{subsec:selection}
Type and eligibility rules first produce $\Aelig(x)\subseteq\CalA(x)$. A policy can request combinations such as ensured counter-oriented candidates, potential candidates for review, or super-oriented candidates satisfying a minimum uncertainty reduction. Eligibility then enforces domain requirements represented in the records. If no candidate satisfies the rules, the output is empty, and the selector does not relax constraints silently.

For ranking, let $\widetilde C_j$ and $\widetilde U_j$ denote coordinates on $[0,1]$. Values already on a suitable common scale can be used directly. Otherwise, the reference procedure applies min--max transformations over the eligible set and fixes them before screening. Scaling affects ranking only: uncertainty change, interval position, and directional typing use the original coordinates. The scaling choice is part of the policy because equal numerical weights do not imply equal substantive importance.

Optional screening retains the union $F=F^+\cup F^-$, where $F^+$ contains candidates nondominated for increasing confidence and decreasing uncertainty, and $F^-$ contains candidates nondominated for decreasing confidence and decreasing uncertainty. The union preserves candidates relevant to either confidence direction. \ref{app:properties} proves preservation of the optimal score under fixed ranking coordinates. The complete set of endpoint ties and the original full-set top-$k$ shortlist need not survive, and applications requiring those outputs should omit screening or specify compatible tie handling.

A signed weight $w\in[-1,1]$ specifies the reference confidence--uncertainty preference:
\begin{align}
D_w(a_j)&=\begin{cases}
\widetilde C_j,&w\geq0,\\
1-\widetilde C_j,&w<0,
\end{cases}\label{eq:direction}\\
S_w(a_j)&=(1-|w|)(1-\widetilde U_j)+|w|D_w(a_j).\label{eq:score}
\end{align}
Thus, $w=0$ minimises uncertainty, $w=1$ maximises confidence, and $w=-1$ minimises confidence. Intermediate magnitudes express a compromise, where the sign identifies the direction. In multiclass classification, decreasing confidence for a specified class does not identify which competing class should be preferred.

For the four cancer candidates, $w=-0.5$ selects $a_1$, $w=0$ ties $a_1$ and $a_3$, and $w=0.5$ selects $a_3$. For positive $w$,
\[
S_w(a_3)=0.90-0.05w,\qquad S_w(a_4)=0.80+0.10w,
\]
so the preferred same-direction candidate switches at $w=2/3$. At $w=0.8$, uncertainty still contributes, but $a_4$ is preferred because the policy accepts its larger uncertainty in exchange for higher confidence. A potential-only request retains $a_2$. Adding a prohibition on changing age produces an empty result. Bidirectional screening of all four candidates removes $a_2$ and retains $a_1$, $a_3$, and $a_4$.

\begin{figure}[t]
\centering
\begin{tikzpicture}
\begin{axis}[
  width=0.88\linewidth,
  height=5.8cm,
  xmin=0,xmax=1,
  ymin=0.2,ymax=1.0,
  xlabel={Positive policy weight $w$},
  ylabel={Reference score $S_w$},
  grid=major,
  legend style={at={(0.5,1.02)},anchor=south,legend columns=4,font=\small},
  tick label style={font=\small},
  label style={font=\small}]
\addplot[black!55,densely dotted,domain=0:1,samples=50]{0.9-0.65*x}; \addlegendentry{$a_1$}
\addplot[black!55,dashdotted,domain=0:1,samples=50]{0.7-0.4*x}; \addlegendentry{$a_2$}
\addplot[blue!70!black,very thick,domain=0:1,samples=50]{0.9-0.05*x}; \addlegendentry{$a_3$}
\addplot[purple!75!black,very thick,dashed,domain=0:1,samples=50]{0.8+0.1*x}; \addlegendentry{$a_4$}
\addplot[black,dashed] coordinates {(0.6666667,0.2) (0.6666667,1.0)};
\node[anchor=south west,font=\small] at (axis cs:0.67,0.27) {$w=2/3$};
\end{axis}
\end{tikzpicture}
\caption{Exact score paths for the four-candidate example when higher confidence is preferred. The upper envelope selects $a_3$ for $0<w<2/3$ and $a_4$ for $w>2/3$; $a_1$ and $a_3$ tie at $w=0$.}
\label{fig:weight-switch}
\end{figure}

\begin{algorithm}[t]
\footnotesize
\caption{Post-generation uncertainty-aware selection}\label{alg:selection}
\begin{algorithmic}[1]
\Require Candidate set $\CalA(x)$, factual reference, policy, shortlist size $k\geq1$
\State Assign supported candidate properties from the available coordinates and metadata.
\State Apply type and eligibility rules to obtain $\Aelig(x)$.
\If{$\Aelig(x)=\emptyset$}
  \State \Return empty result with the active constraints and missing-metadata information.
\EndIf
\State Fix ranking transformations using $\Aelig(x)$.
\State Set $B\gets\Aelig(x)$, or $B\gets F^+\cup F^-$ if screening is requested.
\State Compute $S_w$ on $B$ using the fixed transformations.
\State Rank records using the declared tie rule while retaining tie information.
\State \Return the selected record and the first $\min(k,|B|)$ ranked records, or another requested view.
\end{algorithmic}
\end{algorithm}

The selector returns complete records so that the prediction, uncertainty, rule conditions, and eligibility information remain inspectable. A policy may instead request the eligible set, the screened set, all maximisers, or a ranked shortlist. Records with identical coordinates can encode different feature conditions and need not be interchangeable. Presentation and any subsequent action remain application decisions.

\subsection{Compatibility requirements}\label{subsec:compatibility}
The same selection operations can be applied to generators whose outputs satisfy the requirements in Section~\ref{subsec:interface}. Each mapping retains the interpretation of the generator's prediction and uncertainty signals. Table~\ref{tab:mappings} lists the evaluated and proposed mappings.

\begin{table}[t]
\centering
\footnotesize
\caption{Candidate-provider mappings and qualifications. Proposed mappings are interface arguments, not additional transfer experiments.}
\label{tab:mappings}
\begin{tabularx}{\linewidth}{L{3.0cm}Y}
\toprule
Provider & Candidate coordinates and qualification \\
\midrule
Calibrated Explanations & Task-specific calibrated estimate and associated interval width: main evaluated instantiation. \\
CLUE / $\delta$-CLUE & Decision-relevant prediction and the generator's uncertainty signal, such as predictive entropy or variance: $\delta$-CLUE is evaluated in a supporting transfer check. \\
QUCE & Prediction and a candidate-comparable summary of path uncertainty: path semantics must remain explicit. \\
Multi-objective or robust recourse & Compatible when records expose a decision-relevant prediction and an explicit uncertainty or sensitivity signal: ordinary validity and proximity objectives alone do not establish the mapping. \\
\bottomrule
\end{tabularx}
\end{table}

A provider without a factual uncertainty reference can support uncertainty ranking but not ensured typing. A provider without interval bounds can support ranking but not interval-based potential typing. Pooling outputs from different generators requires a common interpretation of the coordinates, but separate min--max normalisation of unrelated measures is not sufficient.

\section{Instantiation with Calibrated Explanations}\label{sec:ce}
Calibrated Explanations provides local factual and alternative explanations with calibrated predictive information for classification and regression \cite{lofstrom2024calibrated,lofstrom2025calibrated}, including multiclass and conditional extensions \cite{lofstrom2024ce_multiclass,lofstrom2024ce_conditional}. For an alternative $x'$, let $\varphi(x')$ be its task-specific estimate and $[\varphi_{\mathrm{low}}(x'),\varphi_{\mathrm{high}}(x')]$ the associated interval. The present mapping is
\[
C(x')=\varphi(x'),\qquad
U(x')=\varphi_{\mathrm{high}}(x')-\varphi_{\mathrm{low}}(x').
\]

\begin{table}[t]
\centering
\footnotesize
\caption{Task-specific interpretation of the CE mapping.}
\label{tab:ce-mapping}
\begin{tabularx}{\linewidth}{L{2.5cm}L{3.8cm}Y}
\toprule
Task & Prediction coordinate & Uncertainty and directional interpretation \\
\midrule
Binary classification & Calibrated support for the decision-relevant class & Width of the associated Venn--Abers output; boundary-relative types when a decision threshold is defined. \\
Multiclass classification & Calibrated support for a specified target class & Width for that class; decreasing target support does not select a particular competing class. \\
Thresholded regression & Calibrated estimate for the supplied threshold event & Width of the event-specific bounds; threshold is part of the task definition. \\
Plain regression & Numeric prediction & Width of a conformal prediction interval; direction is relative to the factual value or an application reference. \\
\bottomrule
\end{tabularx}
\end{table}

These outputs are not statistically identical objects. Venn--Abers probability outputs should not be assigned the same coverage interpretation as conformal prediction intervals. Interval width is used as an operational within-case uncertainty signal, not as proof of correctness or a universal decomposition of epistemic and aleatoric uncertainty. The selection layer introduces no new marginal, conditional, or post-selection validity guarantee. Any guarantee depends on the assumptions of the underlying calibration or conformal method for the objects being evaluated.

CE produces single-feature alternatives by changing numerical or categorical rule conditions and recalculating candidate estimates and bounds. It also forms conjunctive alternatives by combining feature conditions and obtaining candidate-specific outputs. The selector handles both formats identically once the records are available. Conjunctions can express joint changes while enlarging the set requiring selection. Plausibility and feasibility still depend on the generator and application.

The CE API method \texttt{ensured()} uses the non-increase criterion $U(x')\leq U(x)$. We therefore describe the corresponding empirical counts as uncertainty-non-increasing and reserve ensured for strict reduction. Equal widths do not demonstrate uncertainty reduction.

\section{Empirical Evaluation}\label{sec:evaluation}
The worked example illustrates how candidate properties, eligibility rules, and ranking preferences jointly determine selection, while \ref{app:properties} establishes what optional screening preserves. The empirical evaluation complements these results by examining candidate multiplicity, policy sensitivity within fixed candidate sets, and compatibility with a second uncertainty-aware generator.

\subsection{Datasets and protocol}\label{subsec:protocol}
The main benchmark comprises 41 datasets: five binary classification, 12 multiclass classification, and 24 regression datasets. They contain 768--20,640 instances and 2--40 features. \ref{app:datasets} lists the datasets. Source metadata, preprocessing, scripts, and result artefacts are provided in the reproducibility repository.

All models are random forests with random seed 42: classifiers for classification and regressors for regression. Each dataset supplies a held-out test set of up to 100 instances, and classification splits are stratified. The remaining observations are divided into a proper training set and a calibration pool. Models are fitted on the proper training set and calibrated using a subset of that pool. The aggregate results below use calibration size 100. Holding the learner family and seed fixed isolates selection behaviour within this protocol, but does not establish stability across resampling or model families.

Each regression dataset is evaluated in plain and thresholded modes. Thresholds use the $25^{th}$, $50^{th}$, and $75^{th}$ percentiles of non-test targets. Plain-regression uncertainty is the width between the $5^{th}$ and $95^{th}$ percentiles of the conformal predictive distribution. Single-feature and conjunctive modes are evaluated separately, with the latter including two-feature conjunctions. Several task settings therefore arise from the same underlying dataset and should not be interpreted as independent datasets.

Multiplicity summaries describe generated alternatives and the subset satisfying $U_j\leq U_0$. Policy disagreement is the unweighted mean of within-block disagreement rates over dataset-setting-mode blocks, using cases for which both policies return a selection. Coordinate differences are computed in the ranking coordinates: original coordinates for classification and thresholded regression, and min-max-normalised coordinates for plain regression. Within each block, median absolute differences are calculated over changed selections. The reported values are means of these medians over blocks with changed selections. Ties in the CE comparisons are resolved by the first maximum in candidate order.

\begin{table}[b!]
\centering
\footnotesize
\caption{Candidate multiplicity. Total is the mean number generated. Non-increasing candidates satisfy $U_j\leq U_0$. Conjunctive mode includes two-feature conjunctions.}
\label{tab:multiplicity}
\begin{tabular}{lrrr}
\toprule
Task and mode & Total & Non-increasing & Share (\%)\\
\midrule
Binary, single &18.12&6.48&35.8\\
Binary, conjunctive &69.26&22.77&32.9\\
Multiclass, single &21.75&9.97&45.8\\
Multiclass, conjunctive &69.53&31.59&45.4\\
Regression 25th threshold, single &11.82&5.59&47.3\\
Regression 25th threshold, conjunctive &29.90&13.66&45.7\\
Regression 50th threshold, single &11.57&5.57&48.1\\
Regression 50th threshold, conjunctive &29.48&13.39&45.4\\
Regression 75th threshold, single &11.99&5.95&49.6\\
Regression 75th threshold, conjunctive &30.78&14.91&48.4\\
Plain regression, single &13.30&4.38&32.9\\
Plain regression, conjunctive &32.38&9.57&29.6\\
\bottomrule
\end{tabular}
\end{table}

\subsection{Candidate multiplicity}\label{subsec:multiplicity}
Table~\ref{tab:multiplicity} reports mean generated-set sizes and the uncertainty-non-increasing subset. Single-feature means range from $11.57$ to $21.75$. Including conjunctions increases them to $29.48-69.53$. Uncertainty-non-increasing candidates account for $29.6-49.6\%$ of generated alternatives across the reported settings. Generation therefore regularly leaves collections that require organisation, including alternatives whose uncertainty is no higher than that of the factual reference.

\begin{table}[h!]
\centering
\footnotesize
\caption{Policy disagreement averaged over dataset-setting-mode blocks. Coordinate differences are means of blockwise medians over changed selections, using the coordinates employed for ranking.}
\label{tab:disagreement}
\begin{tabularx}{\linewidth}{Yccc}
\toprule
Policy comparison & Disagree & Mean med. $|\Delta C|$ & Mean med. $|\Delta U|$\\
\midrule
Equal weight vs confidence only ($0.5$ vs $1$) &28.7\%&0.13&0.46\\
Confidence decrease vs increase ($-0.5$ vs $0.5$) &85.5\%&0.65&0.18\\
Equal weight vs uncertainty only ($0.5$ vs $0$) &51.9\%&0.53&0.18\\
Uncertainty only vs confidence only ($0$ vs $1$) &66.4\%&0.50&0.47\\
\bottomrule
\end{tabularx}
\end{table}

\subsection{Policy sensitivity}\label{subsec:sensitivity}
Table~\ref{tab:disagreement} compares selections from fixed candidate sets. Equal-weight ranking ($w=0.5$) and confidence-only ranking ($w=1$) favour the same confidence direction but yield a mean disagreement rate of $28.7\%$ across dataset-setting-mode blocks. For changed selections, the mean blockwise median absolute differences are $0.13$ in the prediction coordinate and $0.46$ in uncertainty. Incorporating uncertainty therefore changes which candidate is selected even when the preferred prediction direction remains fixed.

Reversing confidence direction at equal absolute weight yields a mean disagreement rate of $85.5\%$. The corresponding rates are $51.9\%$ for equal-weight versus uncertainty-only selection and $66.4\%$ for uncertainty-only versus confidence-only selection. Together, these comparisons quantify how policy preferences affect selection within fixed candidate sets.

\begin{figure}[t]
\centering
\begin{subfigure}[t]{0.48\linewidth}
  \centering\includegraphics[width=\linewidth]{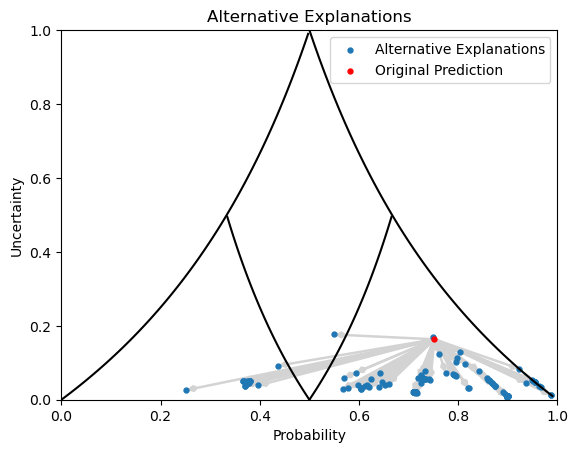}
  \caption{All generated alternatives.}
\end{subfigure}\hfill
\begin{subfigure}[t]{0.48\linewidth}
  \centering\includegraphics[width=\linewidth]{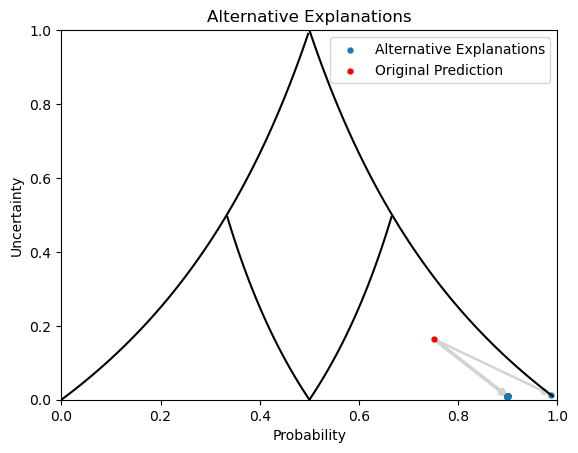}
  \caption{Uncertainty only, $w=0$.}
\end{subfigure}

\begin{subfigure}[t]{0.48\linewidth}
  \centering\includegraphics[width=\linewidth]{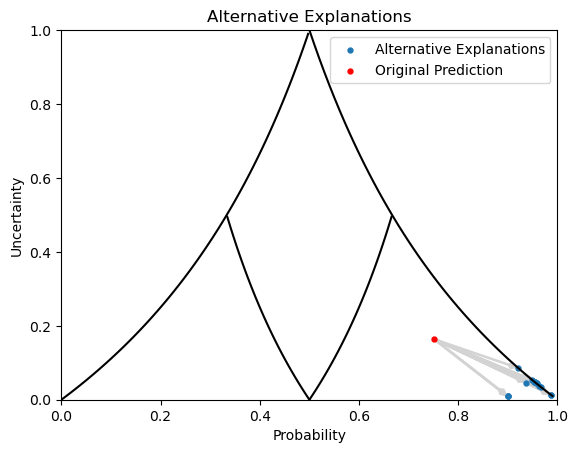}
  \caption{Confidence increasing, $w=1$.}
\end{subfigure}\hfill
\begin{subfigure}[t]{0.48\linewidth}
  \centering\includegraphics[width=\linewidth]{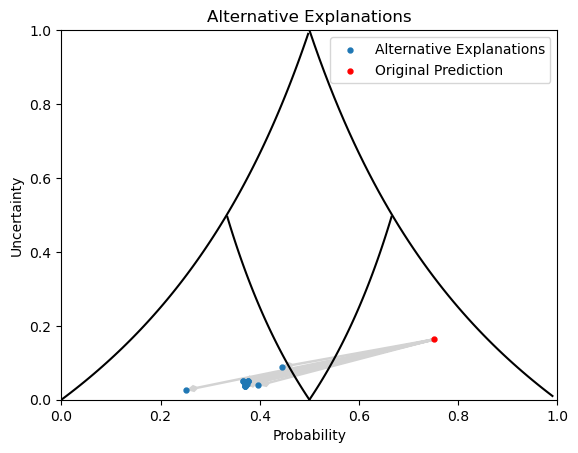}
  \caption{Equal-weight confidence decreasing, $w=-0.5$.}
\end{subfigure}
\caption{Policy sensitivity for a fixed thresholded California Housing candidate set. The panels show the generated set and the top ten records under three policies.}
\label{fig:housing}
\end{figure}

Figure~\ref{fig:housing} provides a candidate-level example from thresholded California Housing. The confidence coordinate represents support for the event $y\leq500$ in this explicitly illustrative setting. The factual case has interval-width uncertainty of approximately $0.18$. Uncertainty-only ranking selects the narrowest candidates. Confidence-only ranking favours stronger support for the event, and an equal-weight confidence-decreasing policy shifts the shortlist toward lower event confidence while retaining a preference for low uncertainty. Several records overlap geometrically despite encoding different rules, which is why selected outputs retain record metadata rather than deduplicating solely by coordinates.

Implementation checks recover uncertainty-only ordering at $w=0$ and the intended confidence ordering at $w=\pm1$. \ref{app:properties} gives the exact optimum-preservation result for optional screening. That result does not imply preservation of every endpoint tie or the original full-set top-$k$ shortlist.

\subsection{Supporting transfer check with \texorpdfstring{$\delta$-CLUE}{delta-CLUE}}\label{subsec:delta}
We apply $\delta$-CLUE \cite{ley2021delta} to the five binary datasets using a neural classifier with Monte Carlo dropout and a variational autoencoder. Each test case uses 20 search starts. Confidence is the probability assigned to the class predicted for the factual case, and uncertainty is predictive entropy divided by $\log 2$. Only candidates with reduced entropy are retained, and duplicates are removed before ranking. The fixed candidate sets are ranked using $w\in\{-1,-0.5,0,0.5,1\}$ without further normalisation. Policy disagreement is calculated over cases where both policies return a candidate.

At least one candidate is obtained for $85.2\%$ of evaluated cases, $84.8\%$ contain multiple candidates, and the mean set size is $16.08$ alternatives per test case. Averaged across the five datasets, top-selection disagreement is $11.3\%$ for $w=-0.5$ versus $w=0.5$ and $12.7\%$ for uncertainty-only versus confidence-only selection. The rates are lower than in the CE study. A more restricted candidate geometry is one possible explanation, but the experiment does not isolate the cause or compare generator quality. It demonstrates compatibility of the basic interface with a second generator. The magnitude and interpretation of policy sensitivity remain dependent on the generator, prediction model, uncertainty statistic, and acceptance rules.

\section{Discussion}\label{sec:discussion}
The framework makes the selection policy an explicit part of explanation use. Candidate properties describe the alternatives produced by the generator, eligibility rules express application constraints, and ranking reflects the current explanatory purpose. An organisation can therefore state which alternatives are admissible and how confidence and uncertainty should influence their priority. Once the candidates have been generated, these choices can be examined and revised using the same set. This makes it possible to trace a selected explanation to both its properties and the policy under which it was chosen.

The worked example shows why uncertainty reduction alone is insufficient to identify an explanatory purpose. Both $a_1$ and $a_3$ are strictly less uncertain than the factual case, yet one challenges the threshold classification, and the other reinforces it. Candidate $a_4$ shows that two alternatives can move in the same confidence direction while trading confidence against uncertainty. Candidate $a_2$ shows that a useful account of model behaviour can still be inadmissible for action-oriented use. These distinctions are especially important when a system presents only one explanation: the selected record should not be portrayed as uniquely supported independently of the policy that produced it.

Selection is conditional on the generated candidate set and the quality and interpretation of its prediction and uncertainty signals. Claims about causal interventions or feasible actions require corresponding evidence and metadata. Within-set normalisation also makes rankings sensitive to changes in the eligible set; fixed external scales may be preferable when stability across such changes is required. Since linear scalarisation can leave some nondominated candidates unselected, applications may retain the frontier or a shortlist to examine a wider range of compromises.

The experiments establish candidate multiplicity and policy sensitivity within the evaluated settings. They do not establish improved human or organisational decisions. The CE study uses one learner family and one random seed, and its aggregate counts concern generated candidates. The proportion of cases retaining multiple distinct eligible candidates after domain restrictions remains to be evaluated. The $\delta$-CLUE study covers five binary datasets under a specified model, uncertainty measure, and acceptance procedure. Further evaluation should examine repeated splits, calibration sizes, dataset-level variation, alternative selectors on identical candidate sets, and the effects of explanation selection on human decision-making.

For deployment, the framework suggests three minimum practices. First, store complete candidate records and the factual reference rather than only a display order. Second, record the policy, transformations, constraints, and tie handling used for each selection. Third, preserve an empty result or visible shortlist when the available candidates do not justify a unique choice. These practices make uncertainty-aware explanation use more inspectable without claiming that transparency alone resolves normative disagreements about whose preferences should govern.

\section{Conclusion}\label{sec:conclusion}
This paper presented a framework for selecting uncertainty-aware explanations according to explicit application policies. It combines changes in prediction and uncertainty with threshold ambiguity, eligibility rules, and policy-aware ranking. The same generated candidate set can therefore support different explanatory questions while retaining the information needed to examine each selection. Experiments across 41 datasets show that incorporating uncertainty changes the selected explanation even when the preferred confidence direction remains fixed. A supporting $\delta$-CLUE experiment demonstrates use with a second generator. The framework makes the role of policy visible in deciding which explanations a system presents and prioritises.

\section*{Acknowledgements}
Helena Löfström acknowledges support from the Kamprad Family Foundation for the project \textit{Förbättrad livskvalitet för äldre genom testning för allvarliga sjukdomar}, grant reference number 20250309. Tuwe Löfström and Johan Hallberg Szabadváry acknowledge the Swedish Knowledge Foundation for financially supporting the research and education environment Knowledge Intensive Product Realisation SPARK at Jönköping University, Sweden, through PREMACOP, grant no. 20220187.

\section*{CRediT authorship contribution statement}
Helena Löfström: Conceptualisation, Methodology, Software, Formal analysis, Writing -- original draft, Writing -- review and editing, Supervision. Tuwe Löfström: Software, Validation, Data curation, Writing -- original draft, Writing -- review and editing. Johan Hallberg Szabadváry: Investigation, Writing -- review and editing.

\section*{Declaration of competing interest}
The authors declare that they have no known competing financial interests or personal relationships that could have appeared to influence the work reported in this paper.

\section*{Data and code availability}
The benchmark datasets used in this study are publicly available. Dataset names, task types, instance counts, and feature counts are reported in \ref{app:datasets}. Dataset-source metadata, preprocessing scripts, experiment scripts, result artefacts, and code required to reproduce the reported diagnostics are available in the \href{https://github.com/Moffran/ENSURED-explanations}{reproducibility repository}.

\appendix
\section{Reference Ranking and Screening Properties}\label{app:properties}
\subsection{Normalisation}
For coordinate $z_j$ over a nonempty eligible set $A$, the reference transformation is
\[
\widetilde z_j=\begin{cases}
\displaystyle\frac{z_j-\min_{a_i\in A}z_i}{\max_{a_i\in A}z_i-\min_{a_i\in A}z_i},&\max z_i>\min z_i,\\[6pt]
0,&\text{otherwise}.
\end{cases}
\]
A constant coordinate contributes equally to every candidate and cannot affect ordering. Transformations are computed on the eligible set and held fixed during screening and ranking. They do not change uncertainty comparisons with the factual case or candidate-type assignments. Scores are local to the candidate set and should not be compared across cases. Adding or removing eligible candidates can change the coordinate ranges and thereby the ranking of remaining candidates.

\subsection{Monotonicity and directional dominance}
\begin{proposition}\label{prop:dominance}
For fixed ranking coordinates, the score in Eq.~\eqref{eq:score} is strictly decreasing in uncertainty when confidence is fixed and $|w|<1$. At fixed uncertainty, it is strictly increasing in confidence for $w>0$ and strictly decreasing for $w<0$. Its endpoints are
\[
S_0=1-\widetilde U,\qquad S_1=\widetilde C,\qquad S_{-1}=1-\widetilde C.
\]
For $0<|w|<1$, if $a_b$ is no worse than $a_a$ in policy-oriented confidence and uncertainty, and strictly better in at least one, then $S_w(a_b)>S_w(a_a)$.
\end{proposition}
\begin{proof}
The coefficient of $1-\widetilde U$ is positive for $|w|<1$, and the coefficient of $D_w$ is positive for $w\neq0$. The definition of $D_w$ establishes the confidence direction. Under directional dominance, both score components are no worse for $a_b$ and one is strictly better. Both coefficients are positive for interior nonzero weights, so the score is strictly greater.
\end{proof}

\subsection{What optional screening preserves}
Let $F^+$ and $F^-$ be the nondominated subsets of the finite eligible set for $(\widetilde C,-\widetilde U)$ and $(-\widetilde C,-\widetilde U)$, respectively. With $F=F^+\cup F^-$ and fixed ranking coordinates,
\[
\max_{a\in A} S_w(a)=\max_{a\in F} S_w(a),\qquad w\in[-1,1].
\]
For $0<w<1$, every maximiser must belong to $F^+$ by Proposition~\ref{prop:dominance}; for $-1<w<0$, the analogous result holds for $F^-$. At $w=1$, a maximum-confidence candidate with minimum uncertainty among tied maximum-confidence candidates is retained in $F^+$. The symmetric argument applies at $w=-1$. At $w=0$, a minimum-uncertainty candidate with maximum confidence among tied minimum-uncertainty candidates belongs to $F^+$. Hence at least one optimum survives at each endpoint.

The complete set of endpoint ties need not survive, nor does the full-set top-$k$ shortlist generally survive for $k>1$. For example, candidates with coordinates $(0.8,0.1)$, $(0.5,0.2)$, and $(0.2,0.1)$ score 0.85, 0.65, and 0.55 at $w=0.5$. The middle candidate ranks second but is dominated in both directional problems and is removed from $F$. A top-$k$ request must therefore specify whether it concerns the eligible set or the screened set. Preserving endpoint tie identities requires a compatible tie rule or omission of screening.

\subsection{Limits of the reference scalarisation}
Varying $w$ need not select every nondominated candidate. Linear scalarisation can leave unsupported candidates on a nonconvex frontier, even when those candidates offer useful compromises. The union frontier and weighted ranking therefore serve different purposes: the first retains efficient alternatives in either confidence direction, while the second encodes one numerical preference. Neither supplies a universal utility function or substitutes for domain-specific eligibility and policy justification.

\setcounter{table}{0}
\section{Benchmark Datasets}\label{app:datasets}
\begin{table}[htbp]
\centering
\scriptsize
\setlength{\tabcolsep}{2pt}
\renewcommand{\arraystretch}{1.05}
\caption{Datasets used in the diagnostic evaluation. Columns are repeated three times to fit the table: DS = dataset, t = task code (b = binary, m = multiclass, r = regression), n = instances, f = features.}
\label{tab:datasets}
\begin{tabular}{lccc|lccc|lccc}
\toprule
DS & t & n & f & DS & t & n & f & DS & t & n & f\\
\midrule
diabetes & b & 768 & 8 & german & b & 955 & 27 & kc1 & b & 1192 & 21\\
pc4 & b & 1343 & 37 & ttt & b & 958 & 27 & cars & m & 1728 & 6\\
cmc & m & 1473 & 9 & cool & m & 768 & 8 & heat & m & 768 & 8\\
image & m & 2310 & 19 & steel & m & 1941 & 27 & vehicle & m & 846 & 18\\
vowel & m & 990 & 11 & wave & m & 5000 & 40 & wineR & m & 1599 & 11\\
wineW & m & 4898 & 11 & yeast & m & 1484 & 8 & abalone & r & 4177 & 8\\
anacalt & r & 4052 & 7 & bank8fh & r & 8192 & 8 & bank8fm & r & 8192 & 8\\
bank8nh & r & 8192 & 8 & bank8nm & r & 8192 & 8 & comp & r & 8192 & 12\\
concrete & r & 1030 & 8 & cooling & r & 768 & 8 & deltaA & r & 7129 & 5\\
deltaE & r & 9517 & 6 & friedm & r & 1200 & 5 & heating & r & 768 & 8\\
housing & r & 20640 & 9 & kin8fh & r & 8192 & 8 & kin8fm & r & 8192 & 8\\
kin8nh & r & 8192 & 8 & kin8nm & r & 8192 & 8 & laser & r & 993 & 4\\
mg & r & 1385 & 6 & mortgage & r & 1048 & 15 & plastic & r & 1649 & 2\\
puma8fh & r & 8192 & 8 & puma8fm & r & 8192 & 8 & & & &\\
\bottomrule
\end{tabular}
\end{table}

\section{Geometry of Interval-Valued Probability Records}\label{app:geometry}
The confidence--uncertainty geometry depends on the definition of the point coordinate. Let $0\leq\ell\leq h\leq1$ and $U=h-\ell$. These algebraic constraints describe probability-valued records; they are not coverage guarantees and do not apply to arbitrary uncertainty signals or unscaled regression predictions.

\paragraph{Midpoint coordinate}
For $C=(\ell+h)/2$, the interval is recovered as
\[
\ell=C-U/2,\qquad h=C+U/2,
\qquad 0\leq U\leq2\min\{C,1-C\}.
\]
The feasible region is triangular. An interval contains a supplied probability boundary $\tau\in(0,1)$ exactly when $U\geq2|C-\tau|$, subject to feasibility. Below that boundary the interval lies wholly on one side of $\tau$.

\paragraph{Single-pair regularised coordinate}
For the probability-pair transformation $C=h/(1-\ell+h)=h/(1+U)$,
\[
h=C(1+U),\qquad \ell=C-(1-C)U.
\]
For $0<C<1$, feasibility becomes
\[
0\leq U\leq\min\left\{\frac{C}{1-C},\frac{1-C}{C}\right\};
\]
at $C=0$ or $C=1$, only $U=0$ is feasible. The interval contains $\tau$ precisely when
\[
U\geq
\begin{cases}
(\tau-C)/C,&0<C<\tau,\\
0,&C=\tau,\\
(C-\tau)/(1-C),&\tau<C<1,
\end{cases}
\]
again subject to feasibility. Thus, changing the point-coordinate convention changes both the outer feasible boundary and the location of interval-ambiguity regions, even for the same interval widths.

\begin{figure}[htbp]
\centering
\begin{subfigure}[t]{0.48\linewidth}
\centering
\begin{tikzpicture}
\begin{axis}[width=\linewidth,height=5.2cm,xmin=0,xmax=1,ymin=0,ymax=1,
 xlabel={$C$},ylabel={$U$},tick label style={font=\scriptsize},label style={font=\small},
 xtick={0,0.5,1},ytick={0,0.5,1},axis lines=left]
\addplot[black,thick] coordinates {(0,0) (0.5,1) (1,0)};
\addplot[blue!70!black,thick,dashed] coordinates {(0.25,0.5) (0.5,0) (0.75,0.5)};
\node[font=\scriptsize] at (axis cs:0.5,0.7) {potential};
\end{axis}
\end{tikzpicture}
\caption{Midpoint: $C=(\ell+h)/2$.}
\end{subfigure}\hfill
\begin{subfigure}[t]{0.48\linewidth}
\centering
\begin{tikzpicture}
\begin{axis}[width=\linewidth,height=5.2cm,xmin=0,xmax=1,ymin=0,ymax=1,
 xlabel={$C$},ylabel={$U$},tick label style={font=\scriptsize},label style={font=\small},
 xtick={0,0.5,1},ytick={0,0.5,1},axis lines=left]
\addplot[black,thick,domain=0:1,samples=100] ({x/(1+x)},{x});
\addplot[black,thick,domain=0:1,samples=100] ({1/(1+x)},{x});
\addplot[blue!70!black,thick,dashed,domain=0:0.5,samples=60] ({0.5/(1+x)},{x});
\addplot[blue!70!black,thick,dashed,domain=0:0.5,samples=60] ({(0.5+x)/(1+x)},{x});
\node[font=\scriptsize] at (axis cs:0.5,0.7) {potential};
\end{axis}
\end{tikzpicture}
\caption{Regularised pair: $C=h/(1-\ell+h)$.}
\end{subfigure}
\caption{Feasible confidence--uncertainty regions for probability intervals. Solid lines delimit feasible records; dashed lines separate intervals wholly on one side of $\tau=1/2$ from potential intervals containing it. The feasible region above the dashed boundaries is potential. These are representation boundaries, not Pareto frontiers. The curved guides in Figure~\ref{fig:housing} use the second geometry.}
\label{fig:interval-geometry}
\end{figure}

The worked cancer example uses the midpoint convention; its decision boundary is $\tau=0.35$, not the illustrative $1/2$ used in Figure~\ref{fig:interval-geometry}. Both mappings are invertible when the point-coordinate convention and width definition are known, so bounds can then be reconstructed. A generic provider need not use either mapping; for that reason the selection contract requires actual interval bounds or a declared valid reconstruction. Aggregation across multiple probability pairs need not preserve the single-pair formula.

\section*{Declaration of generative AI and AI-assisted technologies in the manuscript preparation process}
During the preparation of this work, the authors used ChatGPT (OpenAI) to support language editing, structural condensation, and LaTeX drafting. After using this tool, the authors reviewed and edited the content as needed and take full responsibility for the content of the published article.

\bibliographystyle{elsarticle-num}
\bibliography{Final}

\end{document}